\documentclass[authorversion,sigconf,screen]{acmart}
\usepackage{algorithmicx}
\usepackage{algpseudocode}
\usepackage[linesnumbered,ruled,vlined]{algorithm2e}
\usepackage{float}
\usepackage{listings}
\usepackage{tabularx}

\copyrightyear{2026}
\acmYear{2026}
\setcopyright{cc}
\setcctype{by}
\acmConference[LAK 2026]{LAK26: 16th International Learning Analytics and Knowledge Conference}{April 27-May 01, 2026}{Bergen, Norway}
\acmBooktitle{LAK26: 16th International Learning Analytics and Knowledge Conference (LAK 2026), April 27-May 01, 2026, Bergen, Norway}
\acmPrice{}
\acmDOI{10.1145/3785022.3785042}
\acmISBN{979-8-4007-2066-6/2026/04}

\begin{document}

\title[Representation Learning for Open Response Analytics]{Disentangling Learning from Judgment: Representation Learning for Open Response Analytics}

\author{Conrad Borchers}
\orcid{0000-0003-3437-8979}
\affiliation{
    \institution{Carnegie Mellon University}
    \city{Pittsburgh, PA}
    \country{USA}
}
\email{cborcher@cs.cmu.edu}

\author{Manit Patel}
\orcid{0009-0005-9922-8199}
\affiliation{
    \institution{Carnegie Mellon University}
    \city{Pittsburgh, PA}
    \country{USA}
}
\email{manitp@andrew.cmu.edu}

\author{Seiyon M. Lee}
\orcid{0009-0001-9041-3313}
\affiliation{
    \institution{University of Florida}
    \city{Gainesville, FL}
    \country{USA}
}
\email{leeseiyon@ufl.edu}

\author{Anthony F. Botelho}
\orcid{0000-0002-7373-4959}
\affiliation{
    \institution{University of Florida}
    \city{Gainesville, FL}
    \country{USA}
}
\email{abotelho@coe.ufl.edu}

\renewcommand{\shortauthors}{Borchers et al.}

\newcommand{\revision}[1]{\textcolor{black}{#1}}

\begin{abstract}
Open-ended responses are central to learning, yet automated scoring often conflates what students wrote with how teachers grade. We present an analytics-first framework that separates content signals from rater tendencies, making judgments visible and auditable via analytics. Using de-identified ASSISTments mathematics responses, we model teacher histories as dynamic priors and represent text with sentence embeddings. We apply centroid normalization and response–problem embedding differences, and explicitly model teacher effects with priors to reduce problem- and teacher-related confounds. Temporally-validated linear models quantify the contributions of each signal, and model disagreements surface observations for qualitative inspection. \revision{Results show that} teacher priors heavily influence grade predictions; the strongest results arise when priors are combined with content embeddings (AUC$\approx$0.815), while content-only models remain above chance but substantially weaker (AUC$\approx$0.626). Adjusting for rater effects sharpens the selection of features derived from content representations, retaining more informative embedding dimensions and revealing cases where semantic evidence supports understanding as opposed to surface-level differences in how students respond. The contribution presents a practical pipeline that transforms embeddings from mere features into learning analytics for reflection, enabling teachers and researchers to examine where grading practices align (or conflict) with evidence of student reasoning and learning.
\end{abstract}

\begin{CCSXML}
<ccs2012>
   <concept>
       <concept_id>10010147.10010341</concept_id>
       <concept_desc>Computing methodologies~Modeling and simulation</concept_desc>
       <concept_significance>500</concept_significance>
   </concept>
   <concept>
       <concept_id>10010405.10010489</concept_id>
       <concept_desc>Applied computing~Education</concept_desc>
       <concept_significance>300</concept_significance>
   </concept>
</ccs2012>
\end{CCSXML}

\ccsdesc[500]{Computing methodologies~Modeling and simulation}
\ccsdesc[300]{Applied computing~Education}

\keywords{Open-response assessment, Representation learning, Rater effects, Interpretability}

\maketitle

\section{Introduction}

Open-ended student responses provide some of the most informative evidence of learning \cite{butler2018multiple}. They not only indicate whether an answer is correct but also reveal how students reason, justify, and communicate their understanding. Precisely because of this richness, they are difficult to assess at scale: teachers diverge in how they grade the same work \cite{botelho2023leveraging,gurung2022considerate}, and automated scoring risks confounding content with stylistic or contextual cues \cite{andersen2025algorithmic}. These issues create methodological obstacles for valid modeling and practical obstacles for using such work to support learning. Recent research in learning analytics and related fields highlights remedies such as bias- and context-aware modeling that accounts for rater effects and representation-level robustness that reduces sensitivity to surface features \cite{gibson2022,misgna2024survey}. Yet these approaches remain algorithmic and do not \revision{consider the role of} a central stakeholder: teachers.

We propose a complementary, \emph{analytics-first} approach. \revision{Rather than} accepting automated scores as final, systems should render model behavior auditable by linking predictions to each teacher's grading history and revealing where content signals diverge from rater priors. This is critical because teachers may reasonably rely on automated grades for efficiency, which can reduce opportunities to inspect why a score was assigned \cite{xavier2025}. Our contributions are twofold: (a) a deconfounded representation-learning framework that disentangles student content from teacher priors, and (b) early insights into practitioner-facing analytics that expose disagreement patterns, transforming embeddings from passive features into tools for inspecting open-response grading.

\subsection{The Present Study and Contributions to the Field of Learning Analytics}

We examine how response embeddings can \revision{help} disentangle and interpret the \revision{key} signals in open-ended assessments. Embeddings map the semantic structure of student answers into a continuous vector space \cite{reimers2019sentence}, supporting similarity, clustering, and transformation. In learning analytics \revision{research}, they have revealed semantic distinctions in course offerings \cite{pardos2020university} and dialog moves during learning \cite{borchers2025can}. Such properties \revision{also} make embeddings well-suited for producing interpretable accounts of model behavior that illuminate the boundary between student understanding and grading practice.

We build on the premise that student-authored responses and teacher grading behaviors are distinct yet intertwined signals. Conventional predictive models collapse these dimensions, making it unclear whether predictions stem from student content or from teachers' evaluative tendencies. This conflation undermines the validity of predictions and the pedagogical insights they can offer. Research on annotation disagreement shows that variation across raters often reflects meaningful signal rather than noise \cite{davani2022dealing}. Building on this view, we aim to generate analytics that prompt teachers to reflect on their grading biases and examine student response dimensions most relevant to learning. To achieve this, we introduce a representation-learning framework that explicitly disentangles student content from rater priors.

Methodologically, we introduce and benchmark models and embedding adjustments that isolate content-based information from rater influences. Analytically, we demonstrate how these representations reveal patterns of disagreement that support qualitative analysis. Our study takes important steps toward developing learning analytics for reflection, enabling a \revision{deeper} interpretation of \revision{automated} assessments for both researchers and practitioners. \revision{Specifically,} we ask: 

\begin{itemize}
    \item \textbf{RQ1:} Do models adjusted for teacher grading bias lead to more accurate model predictions?
    \item \textbf{RQ2:} Does adjusting for teacher grading bias lead to more distinctive signals of features in the embedding space, improving model interpretability? 
    \item \textbf{RQ3:} Once rater effects are separated, \revision{what do} embeddings reveal regarding \revision{patterns} of disagreement that correspond to meaningful categories of reasoning and cognitive strategies?
\end{itemize}

\section{Background and Motivation}

\subsection{Automated Open-Response Grading}

Particularly in mathematics, teachers often use open-ended questions to better assess students' conceptual understanding and problem-solving strategies \cite{butler2018multiple}. Unlike closed-ended items, such as multiple-choice or numeric entry, these responses reveal reasoning and thought processes more explicitly. Yet the diversity of expression makes them difficult and time-consuming to evaluate, slowing feedback to students \cite{botelho2023leveraging,gurung2022considerate}. To address this, research has developed methods for automatically scoring open-ended work, both to help teachers provide feedback more efficiently \cite{botelho2023leveraging, erickson2020automated} and to give students timely guidance that can improve their learning \cite{andersen2025algorithmic}.

Early approaches to open-response grading relied on rule- or structure-based representations. For example, C-Rater decomposed answers into knowledge components mapped to rubric criteria \cite{leacock2003c}. Other methods applied clustering \cite{suzen2020automatic} or bag-of-words and n-gram models \cite{ramesh2022automated}. More recently, researchers have pursued representations that capture deeper semantic relationships in student responses, aiming to reflect conceptual understanding across diverse expressions \cite{haller2022survey}. Word embedding models, such as word2vec, GloVe, and BERT, as well as sentence-level methods like SBERT, have been combined with machine learning algorithms, including random forests, XGBoost, and LSTM networks \cite{erickson2020automated}.

Embedding-based approaches have recently surpassed traditional methods \cite{erickson2020automated}, advancing assessment, feedback, and insights into student cognition. SBERT embeddings, for example, have been used to identify similar responses and generate transferable scores and feedback \cite{botelho2023leveraging}. Large language models have also revealed how dialectal variation, such as African American Vernacular English (AAVE), shapes grading outcomes \cite{Siedahmed2025nonstandard}. While much of this work emphasizes gains in scoring accuracy, less attention has been given to distinguishing between linguistic markers of understanding and features reflecting teacher grading practices. This study addresses that gap by disentangling content signals from rater priors, yielding stronger predictive models and more interpretable analyses \revision{both of and for} teacher evaluation.

\subsection{Teacher Bias in Open-Response Grading} 

While recent advances in automatic scoring hold promise for improving grading and feedback, it is crucial to recognize that the labels used to train and evaluate such models originate in human judgment and may reflect contextual factors beyond the student response itself. Variation in how teachers assess open-ended work affects both model performance and the interpretation of what scores reveal about student knowledge and reasoning. Understanding grading practices is, therefore, essential not only for building more accurate models but also for clarifying what teacher-given scores truly represent about student understanding.

A growing body of work shows that teacher-assigned scores for open-ended mathematics responses reflect not only students' answers but also contextual and heuristic factors that vary within and across teachers \cite{botelho2023leveraging,gurung2022considerate,baral2021improving,meissel2017subjectivity}. In experimental re-grading studies on ASSISTments, \revision{for instance}, teachers displayed substantial intra-rater variation when re-evaluating the same \revision{student} responses and, on average, gave slightly lower scores to anonymized work \cite{gurung2022considerate} \revision{than others}. Large-scale error modeling of automated scoring further reveals strong teacher-level effects, even after controlling for answer features \cite{baral2021improving}. These findings suggest that labels for open responses, although ecologically valid, are influenced by pedagogical choices that extend beyond textual content and persist even under rubric-based scoring \cite{ccetin2017analysis}.

Prior work cautions that inferences about knowledge from open-ended responses may not align with teacher-provided scores \cite{gurung2022considerate,meissel2017subjectivity}. In some cases, scores capture students' conceptual levels, while in others they reflect task completion or effort \revision{invested}. Even after accounting for teachers' grading strategies, additional confounds arise from the writing task itself, which draws on both conceptual and procedural knowledge as well as \revision{students'} writing ability \cite{condor2024predicting,schafer2005resistance}. This study addresses these challenges by developing measures that separate content-based indicators of understanding from stylistic features and grading tendencies influencing teacher scores. By isolating rater effects from the semantic content of responses, we aim to yield a clearer signal of students' conceptual knowledge and reasoning while making grading practices more transparent for researchers and \revision{teachers}.

\subsection{Representation Learning for Learning Analytics}

Representation learning compresses educational artifacts—such as texts, problems, and curricular structures—into continuous vectors that retain task-relevant semantics. In learning analytics, such embeddings support not only prediction but also scalable similarity search, structural discovery through clustering and projection, and comparisons across teachers, tasks, and cohorts that sparse features cannot capture. Curricular and pathway analytics benefit from embeddings trained on enrollment and transcript traces that recover course neighborhoods with semantic fidelity rivaling those of catalog descriptions, enabling institutional maps that enhance advising and recommendations \cite{pardos2020university,xu2024extracting}.

Assessment and feedback research has utilized sentence- and document-level embeddings to scale teacher judgment while maintaining a focus on practice. SBERT-style encoders retrieve nearest-neighbor responses for rubric-aligned scoring and feedback in mathematics, including settings with symbolic or image-based work \cite{botelho2023leveraging,baral2023auto}. Other efforts fine-tune sentence encoders on classroom data to support classifiers while preserving interpretable neighborhoods and prototypes for moderation \cite{pijeira2024evaluating}. Embeddings also advance writing and discourse analytics. Reviews trace a shift from hand-crafted linguistic indices to representation-based approaches that capture argumentation, evidence use, and coherence, while delivering actionable feedback in \revision{user-facing interfaces, such as} dashboards \cite{gibson2022}. As in this study, the value \revision{added from this line of work}  lies less in a single accuracy figure than in transforming unstructured traces into analyzable objects that connect directly to pedagogy.

Yet, representation choices shape auditability. Embeddings may entangle demographic or contextual signals in ways that affect fairness and interpretability; recent work quantifies such leakage and explores augmentation to mitigate it \cite{sha2022bigger}. Adopting an analytics-first stance, we treat embeddings not only as predictive features but as lenses through which \revision{teachers} can \revision{more critically} examine grading practices, inspect clusters, and situate their choices within a shared frame.

\section{Methods and Data}

The data\footnote{Our analyses rely on secondary use of de-identified student records in compliance with a data sharing agreement with the ASSISTments platform. Because of these agreements, the dataset cannot be shared publicly. Interested researchers may request access through ASSISTments' established processes at \text{https://www.assistments.org/e-trials}.} \revision{used} for this study come from open-ended mathematics responses on the ASSISTments platform, distributed as the \texttt{QC23-Complete} dataset. We sampled responses from 10{,}000 students, yielding 1{,}581{,}700 records that included free-text submissions, problem, student, and teacher identifiers, teacher-assigned scores, and timestamps.

We applied several preprocessing steps to focus on responses with explanatory reasoning and to ensure variance for modeling teacher effects. Records without a valid skill code were removed, as were responses lacking alphabetic characters, which usually consisted of numeric or symbolic entries. This reduced the dataset to 193{,}241 records. We then excluded responses shorter than ten words, \revision{for} which inspection showed to be guesses, equations, or multiple-choice style entries, leaving 37{,}814 records. To enable multilevel estimation of teacher effects, we excluded teachers whose graded responses showed no variance in correctness. Following recommendations that random effects require within-cluster variability \cite{snijders2011multilevel}, this filtering left 99 teachers contributing 37{,}456 responses. We also removed submissions consisting solely of embedded images, as these cannot be meaningfully represented in text-based models. The final dataset comprised 30{,}585 responses from 2{,}648 students graded by 99 teachers.

\subsection{Study Context: The ASSISTments Platform}

ASSISTments is designed to support mathematics instruction, practice, and assessment in middle and high school classrooms \cite{heffernan2014assistments}. Teachers can create or assign problem sets, with content drawn from widely used open educational resources (OER) such as Illustrative Math, EngageNY, and Utah Math. Teachers select the content to assign and determine how to effectively integrate it into their instruction. This flexibility produces variation in adoption, ranging from core to supplemental use, for formative practice or assessment, and for homework or classwork, but also reflects the authentic ways teachers approach classroom content. 

ASSISTments provides immediate correctness feedback for closed-ended problems (e.g., multiple-choice, fill-in, or numeric-entry items) and allows teachers to assign open-ended mathematics problems, often word problems or ``explain your reasoning'' items, which are manually scored. Although automated scoring and feedback tools have recently been deployed in the system \cite{botelho2023leveraging}, the data analyzed here were collected before their widespread use. Teachers typically score responses on a 0–4 scale, without being bound to rubrics beyond those they may apply outside of what the system logs.

\subsection{Modeling Framework and Experiments}

Our modeling framework integrates two dimensions. First, teachers differ in grading style \cite{gurung2022considerate}, which we capture through dynamic teacher-level averages of past grades \revision{(or what we refer to as teacher priors)}. Second, we incorporate the cognitive dimension through embeddings of responses and problem prompts. We consider response embeddings alone, differences from prompt embeddings to highlight alignment, and concatenated forms to capture joint context. These representations help disentangle content-driven signals from grading biases and learner strategies.

We trained models with progressively richer feature sets, including teacher priors, student priors, and embedding-based content features, as well as integrated models that combine all of these. To enhance interpretability, we used linear Lasso regression with cross-validated regularization. Evaluation relied on a temporally held-out test set (20\%), reporting regression metrics (mean squared error) and classification metrics (ROC AUC after a median split). This design enables us to quantify the influence of teacher bias and student history while testing whether embeddings provide additional predictive value (RQ1), thereby addressing the challenge of distinguishing between grading tendencies and genuine cognitive strategies (RQ2).

\subsection{Interpretability Methods and Projection}

To complement quantitative evaluation, we developed an interpretability pipeline that projects model predictions into a form suitable for qualitative inspection and thematic analysis. The aim was not only to compare \revision{how models performed,} but to understand why different choices \revision{may have} yielded divergent assessments of the same student work, addressing RQ3.

Based on results for RQ1 and RQ2, we focused on three model variants \revision{showing} central contrasts: \emph{response-only}, \emph{teacher+response}, and \emph{teacher-only}. This isolates the contribution of student content, teacher priors, and their combination. To maximize analytic value, we set a single prediction threshold that produced the greatest divergence across the three models, then filtered the test set to retain only instances with conflicting binary predictions. This process yielded over 1,800 disagreement cases, most of which involved teacher+response and teacher-only predicting correct, while response-only did not ($n \approx 1{,}410$). 

From this pool, we sampled about 300 cases for \revision{an in-depth,  qualitative analysis}. Sampling was guided by a \emph{prototypical score}, defined as the cosine similarity between each response embedding and the centroid of its disagreement cluster. This enabled the selection of both central and extreme examples (50\% of the sample each), ensuring that coders encountered representative cases as well as outliers that highlighted less typical patterns. Each case was exported to a shared spreadsheet, which included the student's response, correctness label, identifiers, and binary predictions from all three models. Disagreement patterns were displayed both in compact numeric form (e.g., \texttt{1-0-1}) and as text.

Two research team members first coded responses independently, then discussed and consolidated their codes into three main categories: conceptual explanation (responses explaining reasoning), procedural explanation (responses listing only computational steps), and unclassifiable responses (ambiguous, unrelated, or unreadable). In a final round, coded responses were grouped by disagreement pattern to analyze associations between content and model predictions. The coders then curated results through discussion.

\revision{Our methodology is further documented in this study's supplemental GitHub repository.\footnote{\url{https://github.com/conradborchers/math-representations-lak26}}}

\section{Results}

The analytic procedures described above directly map onto our three research questions. To investigate RQ1, we compared alternative embedding constructions in models trained and evaluated under the same temporal validation framework, allowing us to quantify which representations most reliably predict teacher-assigned grades. For RQ2, we extended these models with teacher and student priors. We examined how many features remained predictive under the adjusted embedding variants, thereby assessing whether accounting for rater leniency and severity sharpens predictive and interpretive signals. For RQ3, we group the student responses into three distinct categories: conceptual explanation, procedural explanation, and insufficient information. We analyze how model predictions differ for these groups, aiming to identify which signals in the response content and priors are captured by the three models by focusing on disagreement patterns for each group.

\subsection{RQ1: Do models adjusted for teacher bias lead to better predictions?}

Table~\ref{tab:results} reports model performance on the temporally held-out test set using mean squared error (MSE) and AUC. Across all specifications, models that incorporate teacher priors achieve the highest predictive accuracy. The strongest performance is achieved when response embeddings are combined with teacher priors, yielding an AUC of 0.815 (95\% CI [0.804, 0.827]) with an MSE of 0.0619. This slightly outperforms both the centroid-normalized variant (AUC = 0.813, MSE = 0.0626) and the teacher–prior–only baseline (AUC = 0.799, MSE = 0.0628). The small differences in MSE among these top three models, all within overlapping confidence intervals, indicate that improvements are most visible in discriminative capacity (AUC) rather than raw error reduction.

By contrast, models that exclude priors and rely solely on response content are substantially weaker. A model based only on student responses yields an AUC of 0.626 (95\% CI [0.611, 0.641]; MSE = 0.0771), which is reliably above chance but falls short of the performance of any teacher-aware model. Combining response and problem embeddings yields nearly identical results (AUC = 0.632; MSE = 0.0772). Problem embeddings in isolation, or differencing-style constructions that subtract prompt vectors, perform only marginally better than random guessing (AUCs in the 0.56–0.62 range).

\begin{table*}[htpb]
\centering
\caption{Predictive performance on the held-out test set, including bootstrapped 95\% confidence intervals.}
\begin{tabular}{lcc}
\hline
Model & MSE [95\% CI] & AUC [95\% CI] \\
\hline
Teacher prior + Response embedding        & 0.0619 [0.0588, 0.0646] & 0.815 [0.804, 0.827] \\
Teacher prior + Response (centroid-normalized) & 0.0626 [0.0598, 0.0654] & 0.813 [0.801, 0.825] \\
Teacher prior only                        & 0.0628 [0.0598, 0.0658] & 0.799 [0.787, 0.812] \\
Teacher prior + Response–Problem difference & 0.0631 [0.0601, 0.0659] & 0.798 [0.786, 0.811] \\
Problem + Response embeddings             & 0.0772 [0.0741, 0.0801] & 0.632 [0.618, 0.648] \\
Response embedding only                   & 0.0771 [0.0741, 0.0800] & 0.626 [0.611, 0.641] \\
Response–Problem difference only          & 0.0774 [0.0743, 0.0802] & 0.617 [0.602, 0.634] \\
Response (centroid-normalized) only       & 0.0804 [0.0774, 0.0833] & 0.590 [0.573, 0.606] \\
Problem embedding only (baseline)                    & 0.0789 [0.0757, 0.0818] & 0.560 [0.542, 0.576] \\
\hline
\end{tabular}
\label{tab:results}
\end{table*}

\subsection{RQ2: Do models adjusted for teacher bias lead to more distinct signals?}

Adjustments for teacher bias \revision{yield} a more distinct and interpretable content signal. In the unadjusted model, the Lasso retains 18 of 384 embedding coordinates (4.7\%), suggesting that strong, low-variance priors suppress many content features. After orthogonalization, 40 of 384 coordinates (10.4\%) remain nonzero. 

Evidence from disagreement patterns on the held-out set reinforces this conclusion. The most frequent configuration is that the teacher-only and teacher+response models predict correctness while the response-only model does not ($n \approx 1{,}410$), highlighting the explanatory strength of rater baselines. Yet cases where only the combined model predicts correctness ($n=126$) or only the response-only model does so ($n=110$) show that, once priors are modeled, semantic content contributes an incremental signal. The rare instances where both content-aware models predict correctness, while the teacher-only model does not ($n=60$), are particularly diagnostic: they identify responses in which content evidence legitimately counterbalances baseline leniency or severity, and thus should be surfaced for practitioner review. We interpret these cases in the next section.

\subsection{RQ3: What do disagreement patterns reveal about the signals captured by different models?}

Our qualitative analysis of disagreement cases sheds light on how content and priors interact, and what this reveals about student reasoning and teacher grading. When the response-only model predicted correctness, disagreements often centered on conceptual reasoning or responses with insufficient information. Teachers frequently awarded partial credit even when answers were ambiguous, as in the case of a student who wrote: ``\textit{100 = the printer, 005p = store bought paper, 025 = office store printing paper}.'' Despite the unclear meaning, both the model and teachers treated it as correct. The response-only model also aligned with teachers when handling ``bad samples'' such as symbolic notation or formatting artifacts, suggesting that embeddings are sensitive to technical inputs that can approximate human judgment.

The prior-only model, by contrast, tended to mark incomplete or irrelevant responses as correct, reflecting how historical performance influenced both human and model assessments. For example, one student described a geometric sequence with ``\textit{1 to 4 to 7 to 10}'' without explanation. Coders judged this insufficient, yet teachers and the model considered it correct. Even off-task remarks such as ``\textit{I don't know I just kinda did it}'' sometimes received credit when student priors were strong, highlighting how outcome-oriented heuristics can overshadow explanatory quality.

The combined model aligned most closely with teachers on procedural answers, particularly those that involved equations or symbolic steps. It also captured many instances of conceptual reasoning, but it occasionally failed when students offered confident but flawed explanations. One student, asked to articulate exponent rules, responded: ``\textit{6\textsuperscript{2} $\times$ 6\textsuperscript{0} it would be 3600}.'' The teacher awarded partial credit despite the incorrect reasoning, while the model also correctly predicted the answer. Such cases demonstrate that integrated models can conflate surface plausibility with genuine conceptual understanding.

Taken together, these disagreement patterns reveal that each signal—response content, rater priors, and their integration—captures different dimensions of the grading process. Content-based models highlight linguistic and conceptual features, prior-based models reflect historical expectations and grading generosity, and combined models approximate the pragmatic compromises teachers often make. Importantly, the rare cases where content models predict correctness against the weight of teacher priors expose responses in which genuine reasoning may be overlooked. These are precisely the situations where an analytics-first approach can make disagreement visible, turning embeddings into diagnostic tools for reflection rather than simply predictive features.

\section{Discussion}

We reframe automated analysis of open responses from a focus on predicting grades to a focus on making the grading process itself intelligible. Our central novelty claim is to separate content signals encoded in response embeddings from rater tendencies captured by teacher priors, then to read their points of convergence and divergence as evidence about what is being modeled. In doing so, we treat open responses as pedagogically meaningful artifacts whose interpretation must remain visible to teachers and researchers \cite{gibson2022,botelho2023leveraging,gurung2022considerate,xavier2025}.

The results for RQ1 indicate that student response embeddings alone convey a real but limited signal regarding teacher-assigned grades. Models that used response embeddings without rater context outperformed chance yet fell short of the performance achieved by teacher-aware models. This pattern is consistent with two complementary interpretations. First, teacher scores encode more than semantic content, reflecting classroom context, expectations, and pragmatic choices that are not present in the text \cite{gurung2022considerate,meissel2017subjectivity}. For instance, the strong predictive performance of teacher leniency may be related to differences in their grading practices and standards, which explain a substantial portion of the variation in grades in our multi-teacher sample. Second, encoders such as SBERT capture only a subset of the cognitive and rhetorical properties teachers reward \cite{reimers2019sentence,erickson2020automated}. In a domain like ratios and proportions, teachers may attend to specific terminologies and reasoning templates; embeddings register some of these regularities but not all, especially when reasoning is implicit or distributed across symbols and prose. We interpret this finding, reflected in the predictive performance of our models, as validation of the idea that learning analytics must develop methods to distinguish between surface-level and content-level grading behavior in order to measure cognitive learner strategies in their responses (e.g., self-explanation). This idea aligns with past work on automated grading, which has found that surface-level features often skew grading \cite{botelho2023leveraging,gurung2022considerate,andersen2025algorithmic}. 

Addressing RQ2, accounting for teacher leniency and severity, sharpened the content signal rather than diluting it. After partialling out rater baselines and problem-level regularities \cite{enders2007centering,zhang2025deconfounding}, more embedding coordinates survived Lasso regularization, and discrimination improved at comparable or lower error. This pattern suggests that priors otherwise absorb variance that overlaps with content features; once that variance is allocated to explicit rater terms, the remaining coefficients concentrate on semantically coherent directions. The effect is methodological and interpretive. Methodologically, the adjusted design yields models that generalize better across teachers because the confounding pathway from rater practice to content proxies is attenuated. Interpretively, the nonzero embedding dimensions are easier to audit because they align with content that remains predictive even when rater baselines are held fixed. For learning analytics, these sharpened signals create opportunities to surface the linguistic and conceptual features of student work that most influence grading. Presenting such dimensions back to teachers can foster reflection and refinement of scoring practices. In future work, we plan to develop teacher-facing analytics that highlight these features to support calibration and alignment with evidence of student understanding.

RQ3 surfaced patterns of disagreement that reveal grading behavior. The projection and coding of disagreements indicate that each signal family captures a distinct facet of grading. Prior-only predictions tended to treat incomplete or thinly supported answers as correct when a student's history had been strong, mirroring the halo-like regularities observed in \revision{prior work on teacher re-grading (e.g., \cite{gurung2022considerate})}. Response-only predictions were most distinctive for concept-laden phrasing and recognizable solution \revision{schema}, as well as for technically formatted inputs that function as plausible evidence of procedure. Combined models tracked many of the pragmatic compromises visible in teacher scoring, where partial reasoning, correct structure with minor execution slips, or confident prose received credit even in the absence of fully articulated explanations \cite{botelho2023leveraging,andersen2025algorithmic}. Rare cases where content-based models judged an answer correct but prior-only models did not are especially useful. They highlight responses where the meaning shows genuine understanding even though the statistical baseline suggests otherwise, making them good candidates for human review. In an analytics-first workflow \revision{as in this work}, these cases are not errors to be suppressed but opportunities for moderation and discussion \cite{davani2022dealing}. In particular, our models could curate counterfactual response pairs where teachers graded differently despite comparable content (e.g., similar values in response dimensions our adjusted model surfaced matter for performance), revealing how surface features shape evaluation and offering material for reflection and calibration.

\subsection{Implications and Future Directions for Learning Analytics}

Our findings demonstrate that, by disentangling response semantics from rater history, models can reveal the provenance of a prediction and clarify the extent to which outcomes reflect student language or prior grading patterns. This enhances interpretability and supports critical engagement with automated judgments \cite{gibson2022,xavier2025}. This work-in-progress study can derive implications for design (although we believe further co-design with teachers will be needed to ensure practical and desirable designs that align with teachers' data preferences and needs \cite{yang2024leveraging}). As initial design directions, predictive dashboards could present the decomposed contributions of rater priors and content similarity neighborhoods, allowing teachers to situate their current decisions against historical tendencies and exemplar cases \cite{gibson2022, botelho2023leveraging}. Disagreements could be made visible as diagnostic objects that inform moderation and rubric alignment, rather than being treated as noise. Because representational choices determine which linguistic and contextual features are legible to the model, they must be considered policy decisions with consequences for fairness and auditability \cite{sha2022bigger}.

Methodologically, this work demonstrates that linear models paired with explicit rater covariates and simple embedding transformations provide a transparent baseline for open-response analytics. Prior work has demonstrated the utility of embeddings for automated scoring \cite{erickson2020automated,botelho2023leveraging}; here, we extend this by showing that explicitly modeling rater effects alongside content embeddings enables the content signal to generalize across teachers and produce interpretable artifacts for qualitative inspection.

\subsection{Limitations and Future Work}

First, this study relies on a single dataset from a single platform, which is dominated by middle school mathematics items on ratios and proportions. The narrow topical range constrains the expressiveness of text-only embeddings, which cannot fully capture strategies, diagrams, or symbolic layout that teachers often consider \cite{erickson2020automated,reimers2019sentence}. Extending analyses to other domains and longer forms of discourse is necessary to assess generalizability. Second, ground-truth labels are shaped by heterogeneous teacher practices. Although rater history was modeled, unobserved factors such as rubrics, instructional goals, or grading conditions remain influential \cite{gurung2022considerate,meissel2017subjectivity}. Moreover, filtering short and image-only responses stabilized modeling but excluded authentic forms of classroom work, biasing the sample toward more verbose students. Other data sets may not require the same level of filtering we applied, which would result in a more representative sample. Third, the modeling design emphasized interpretability through linear sparse estimators. While this yields transparent attributions, it may underfit higher-order interactions. More complex machine learning architectures could improve predictive accuracy, but might obscure the foundations of model decisions. \revision{Fourth, we acknowledge that the embeddings used for analysis may capture biases inherent in language models (e.g., gender bias when assessing student work \cite{du2025benchmarking}). Separating out relevant dimensions of bias in learned representations, including the various potential reasons why teachers exhibit grading biases, is a separate \revision{but a fruitful} line of future work.}

\revision{Finally, we emphasize that the present work is methodological in scope.} Our primary contribution is to learning analytics researchers and system designers rather than directly to classroom practitioners. 
\revision{As such, its primary contribution is more directed toward learning analytics researchers and system designers, rather than directly to classroom practitioners. For our work to achieve the broader impact that motivated it, future research should focus on translating the approaches introduced here into learning analytics capabilities that support instructional use and provide actionable guidance for teachers, including tools that help surface and reflect on potential grading biases.} Our results establish analytic techniques that make rater effects and content signals separable and auditable, laying a foundation for the principled design and co-development of such future tools.

\section{Conclusion}

We recast open-response analytics as the task of making judgment visible. Modeling teacher priors alongside response embeddings reveals that much of the predictive signal originates from the rater context; once this context is explicit, the remaining contribution of the response-based representations becomes easier to interpret and audit. Disagreement patterns between content-only, prior-only, and integrated models serve as diagnostic evidence about why a score was produced, allowing systems to surface cases where the content of a student’s response supports understanding but the predicted score is driven primarily by historical grading patterns. Such cases are natural candidates for teacher review, moderation, or discussion.

Our contribution is a novel pipeline for creating learning analytics that can inform teacher- and researcher-facing tools in future work. This pipeline involves temporal validation with explicit rater baselines and methods for combining these baselines with representation learning for qualitative inspection, which transforms embeddings into tools for reflection. This allows the end-user to decompose predictions, enabling them to interrogate how student reasoning and their evaluation interact. These interrogations will contribute to making teacher- and AI-graded student open responses fairer and more aligned with cognitive strategies that are known to enhance learning, such as self-explanation.

\begin{acks}
\revision{The authors would like to thank the ASSISTments Foundation for the provision and curation of the \texttt{QC23-Complete} dataset. CB wishes to thank Kevin K. Tang for contributions to the data preprocessing code, as well as Vincent Aleven for co-supervision of MP. MP was supported by the Summer Undergraduate Research Apprenticeship (SURA) program at Carnegie Mellon University.}
\end{acks}

\bibliographystyle{ACM-Reference-Format}
\bibliography{main}


\begin{thebibliography}{32}

%%% the Web address.

\ifx \showCODEN    \undefined \def \showCODEN     #1{\unskip}     \fi
\ifx \showISBNx    \undefined \def \showISBNx     #1{\unskip}     \fi
\ifx \showISBNxiii \undefined \def \showISBNxiii  #1{\unskip}     \fi
\ifx \showISSN     \undefined \def \showISSN      #1{\unskip}     \fi
\ifx \showLCCN     \undefined \def \showLCCN      #1{\unskip}     \fi
\ifx \shownote     \undefined \def \shownote      #1{#1}          \fi
\ifx \showarticletitle \undefined \def \showarticletitle #1{#1}   \fi
\ifx \showURL      \undefined \def \showURL       {\relax}        \fi
% invisible to TeX
\providecommand\bibfield[2]{#2}
\providecommand\bibinfo[2]{#2}
\providecommand\natexlab[1]{#1}
\providecommand\showeprint[2][]{arXiv:#2}

\bibitem[Andersen et~al\mbox{.}(2025)]%
        {andersen2025algorithmic}
\bibfield{author}{\bibinfo{person}{Nico Andersen}, \bibinfo{person}{Julia Mang}, \bibinfo{person}{Frank Goldhammer}, {and} \bibinfo{person}{Fabian Zehner}.} \bibinfo{year}{2025}\natexlab{}.
\newblock \showarticletitle{Algorithmic Fairness in Automatic Short Answer Scoring}.
\newblock \bibinfo{journal}{\emph{International Journal of Artificial Intelligence in Education}} (\bibinfo{year}{2025}), \bibinfo{pages}{1--38}.
\newblock


\bibitem[Baral et~al\mbox{.}(2023)]%
        {baral2023auto}
\bibfield{author}{\bibinfo{person}{Sami Baral}, \bibinfo{person}{Anthony Botelho}, \bibinfo{person}{Abhishek Santhanam}, \bibinfo{person}{Ashish Gurung}, \bibinfo{person}{Li Cheng}, {and} \bibinfo{person}{Neil Heffernan}.} \bibinfo{year}{2023}\natexlab{}.
\newblock \showarticletitle{Auto-scoring Student Responses with Images in Mathematics}. In \bibinfo{booktitle}{\emph{Proceedings of the 16th International Conference on Educational Data Mining}}. \bibinfo{pages}{362--369}.
\newblock


\bibitem[Baral et~al\mbox{.}(2021)]%
        {baral2021improving}
\bibfield{author}{\bibinfo{person}{Sami Baral}, \bibinfo{person}{Anthony~F Botelho}, \bibinfo{person}{John~A Erickson}, \bibinfo{person}{Priyanka Benachamardi}, {and} \bibinfo{person}{Neil~T Heffernan}.} \bibinfo{year}{2021}\natexlab{}.
\newblock \showarticletitle{Improving Automated Scoring of Student Open Responses in Mathematics.}
\newblock \bibinfo{journal}{\emph{International Educational Data Mining Society}} (\bibinfo{year}{2021}).
\newblock


\bibitem[Borchers and Shou(2025)]%
        {borchers2025can}
\bibfield{author}{\bibinfo{person}{Conrad Borchers} {and} \bibinfo{person}{Tianze Shou}.} \bibinfo{year}{2025}\natexlab{}.
\newblock \showarticletitle{Can large language models match tutoring system adaptivity? a benchmarking study}. In \bibinfo{booktitle}{\emph{International Conference on Artificial Intelligence in Education}}. Springer, \bibinfo{pages}{407--420}.
\newblock


\bibitem[Botelho et~al\mbox{.}(2023)]%
        {botelho2023leveraging}
\bibfield{author}{\bibinfo{person}{Anthony Botelho}, \bibinfo{person}{Sami Baral}, \bibinfo{person}{John~A Erickson}, \bibinfo{person}{Priyanka Benachamardi}, {and} \bibinfo{person}{Neil~T Heffernan}.} \bibinfo{year}{2023}\natexlab{}.
\newblock \showarticletitle{Leveraging natural language processing to support automated assessment and feedback for student open responses in mathematics}.
\newblock \bibinfo{journal}{\emph{Journal of computer assisted learning}} \bibinfo{volume}{39}, \bibinfo{number}{3} (\bibinfo{year}{2023}), \bibinfo{pages}{823--840}.
\newblock


\bibitem[Butler(2018)]%
        {butler2018multiple}
\bibfield{author}{\bibinfo{person}{Andrew~C Butler}.} \bibinfo{year}{2018}\natexlab{}.
\newblock \showarticletitle{Multiple-choice testing in education: Are the best practices for assessment also good for learning?}
\newblock \bibinfo{journal}{\emph{Journal of Applied Research in Memory and Cognition}} \bibinfo{volume}{7}, \bibinfo{number}{3} (\bibinfo{year}{2018}), \bibinfo{pages}{323--331}.
\newblock


\bibitem[{\c{C}}etin and Ilhan(2017)]%
        {ccetin2017analysis}
\bibfield{author}{\bibinfo{person}{Bayram {\c{C}}etin} {and} \bibinfo{person}{Mustafa Ilhan}.} \bibinfo{year}{2017}\natexlab{}.
\newblock \showarticletitle{An analysis of rater severity and leniency in open-ended mathematic questions rated through standard rubrics and rubrics based on the SOLO taxonomy}.
\newblock \bibinfo{journal}{\emph{Education and Science}} \bibinfo{volume}{42}, \bibinfo{number}{189} (\bibinfo{year}{2017}).
\newblock


\bibitem[Condor(2024)]%
        {condor2024predicting}
\bibfield{author}{\bibinfo{person}{Aubrey Condor}.} \bibinfo{year}{2024}\natexlab{}.
\newblock \showarticletitle{Predicting Short Response Ratings with Non-Content Related Features: A Hierarchical Modeling Approach}. In \bibinfo{booktitle}{\emph{Proceedings of the 18th International Conference of the Learning Sciences-ICLS 2024, pp. 1227-1230}}. International Society of the Learning Sciences.
\newblock


\bibitem[Davani et~al\mbox{.}(2022)]%
        {davani2022dealing}
\bibfield{author}{\bibinfo{person}{Aida~Mostafazadeh Davani}, \bibinfo{person}{Mark D{\'\i}az}, {and} \bibinfo{person}{Vinodkumar Prabhakaran}.} \bibinfo{year}{2022}\natexlab{}.
\newblock \showarticletitle{Dealing with disagreements: Looking beyond the majority vote in subjective annotations}.
\newblock \bibinfo{journal}{\emph{Transactions of the Association for Computational Linguistics}}  \bibinfo{volume}{10} (\bibinfo{year}{2022}), \bibinfo{pages}{92--110}.
\newblock


\bibitem[Du et~al\mbox{.}(2025)]%
        {du2025benchmarking}
\bibfield{author}{\bibinfo{person}{Yishan Du}, \bibinfo{person}{Conrad Borchers}, {and} \bibinfo{person}{Mutlu Cukurova}.} \bibinfo{year}{2025}\natexlab{}.
\newblock \showarticletitle{Benchmarking Educational LLMs with Analytics: A Case Study on Gender Bias in Feedback}.
\newblock \bibinfo{journal}{\emph{arXiv preprint arXiv:2511.08225}} (\bibinfo{year}{2025}).
\newblock


\bibitem[Enders and Tofighi(2007)]%
        {enders2007centering}
\bibfield{author}{\bibinfo{person}{Craig~K Enders} {and} \bibinfo{person}{Davood Tofighi}.} \bibinfo{year}{2007}\natexlab{}.
\newblock \showarticletitle{Centering predictor variables in cross-sectional multilevel models: a new look at an old issue.}
\newblock \bibinfo{journal}{\emph{Psychological methods}} \bibinfo{volume}{12}, \bibinfo{number}{2} (\bibinfo{year}{2007}), \bibinfo{pages}{121}.
\newblock


\bibitem[Erickson et~al\mbox{.}(2020)]%
        {erickson2020automated}
\bibfield{author}{\bibinfo{person}{John~A Erickson}, \bibinfo{person}{Anthony~F Botelho}, \bibinfo{person}{Steven McAteer}, \bibinfo{person}{Ashvini Varatharaj}, {and} \bibinfo{person}{Neil~T Heffernan}.} \bibinfo{year}{2020}\natexlab{}.
\newblock \showarticletitle{The automated grading of student open responses in mathematics}. In \bibinfo{booktitle}{\emph{Proceedings of the tenth international conference on learning analytics \& knowledge}}. \bibinfo{pages}{615--624}.
\newblock


\bibitem[Gibson and Shibani(2022)]%
        {gibson2022}
\bibfield{author}{\bibinfo{person}{Andrew Gibson} {and} \bibinfo{person}{Antonette Shibani}.} \bibinfo{year}{2022}\natexlab{}.
\newblock \showarticletitle{Natural language processing: Writing analytics}.
\newblock In \bibinfo{booktitle}{\emph{Handbook of Learning Analytics--Second edition}}.
\newblock


\bibitem[Gurung et~al\mbox{.}(2022)]%
        {gurung2022considerate}
\bibfield{author}{\bibinfo{person}{Ashish Gurung}, \bibinfo{person}{Anthony Botelho}, \bibinfo{person}{Russell Thompson}, \bibinfo{person}{Adam Sales}, \bibinfo{person}{Sami Baral}, {and} \bibinfo{person}{Neil Heffernan}.} \bibinfo{year}{2022}\natexlab{}.
\newblock \showarticletitle{Considerate, unfair, or just fatigued? examining factors that impact teacher}. In \bibinfo{booktitle}{\emph{Proceedings of the 30th International Conference on Computers in Education. Asia-Pacific Society for Computers in Education}}.
\newblock


\bibitem[Haller et~al\mbox{.}(2022)]%
        {haller2022survey}
\bibfield{author}{\bibinfo{person}{Stefan Haller}, \bibinfo{person}{Adina Aldea}, \bibinfo{person}{Christin Seifert}, {and} \bibinfo{person}{Nicola Strisciuglio}.} \bibinfo{year}{2022}\natexlab{}.
\newblock \showarticletitle{Survey on automated short answer grading with deep learning: from word embeddings to transformers}.
\newblock \bibinfo{journal}{\emph{arXiv preprint arXiv:2204.03503}} (\bibinfo{year}{2022}).
\newblock


\bibitem[Heffernan and Heffernan(2014)]%
        {heffernan2014assistments}
\bibfield{author}{\bibinfo{person}{Neil~T Heffernan} {and} \bibinfo{person}{Cristina~Lindquist Heffernan}.} \bibinfo{year}{2014}\natexlab{}.
\newblock \showarticletitle{The ASSISTments ecosystem: Building a platform that brings scientists and teachers together for minimally invasive research on human learning and teaching}.
\newblock \bibinfo{journal}{\emph{International Journal of Artificial Intelligence in Education}} \bibinfo{volume}{24}, \bibinfo{number}{4} (\bibinfo{year}{2014}), \bibinfo{pages}{470--497}.
\newblock


\bibitem[Leacock and Chodorow(2003)]%
        {leacock2003c}
\bibfield{author}{\bibinfo{person}{Claudia Leacock} {and} \bibinfo{person}{Martin Chodorow}.} \bibinfo{year}{2003}\natexlab{}.
\newblock \showarticletitle{C-rater: Automated scoring of short-answer questions}.
\newblock \bibinfo{journal}{\emph{Computers and the Humanities}} \bibinfo{volume}{37}, \bibinfo{number}{4} (\bibinfo{year}{2003}), \bibinfo{pages}{389--405}.
\newblock


\bibitem[Meissel et~al\mbox{.}(2017)]%
        {meissel2017subjectivity}
\bibfield{author}{\bibinfo{person}{Kane Meissel}, \bibinfo{person}{Frauke Meyer}, \bibinfo{person}{Esther~S Yao}, {and} \bibinfo{person}{Christine~M Rubie-Davies}.} \bibinfo{year}{2017}\natexlab{}.
\newblock \showarticletitle{Subjectivity of teacher judgments: Exploring student characteristics that influence teacher judgments of student ability}.
\newblock \bibinfo{journal}{\emph{Teaching and teacher education}}  \bibinfo{volume}{65} (\bibinfo{year}{2017}), \bibinfo{pages}{48--60}.
\newblock


\bibitem[Misgna et~al\mbox{.}(2024)]%
        {misgna2024survey}
\bibfield{author}{\bibinfo{person}{Haile Misgna}, \bibinfo{person}{Byung-Won On}, \bibinfo{person}{Ingyu Lee}, {and} \bibinfo{person}{Gyu~Sang Choi}.} \bibinfo{year}{2024}\natexlab{}.
\newblock \showarticletitle{A survey on deep learning-based automated essay scoring and feedback generation}.
\newblock \bibinfo{journal}{\emph{Artificial Intelligence Review}} \bibinfo{volume}{58}, \bibinfo{number}{2} (\bibinfo{year}{2024}), \bibinfo{pages}{36}.
\newblock


\bibitem[Pardos and Nam(2020)]%
        {pardos2020university}
\bibfield{author}{\bibinfo{person}{Zachary~A Pardos} {and} \bibinfo{person}{Andrew Joo~Hun Nam}.} \bibinfo{year}{2020}\natexlab{}.
\newblock \showarticletitle{A university map of course knowledge}.
\newblock \bibinfo{journal}{\emph{PloS one}} \bibinfo{volume}{15}, \bibinfo{number}{9} (\bibinfo{year}{2020}), \bibinfo{pages}{e0233207}.
\newblock


\bibitem[Pijeira-D{\'\i}az et~al\mbox{.}(2024)]%
        {pijeira2024evaluating}
\bibfield{author}{\bibinfo{person}{H{\'e}ctor~J Pijeira-D{\'\i}az}, \bibinfo{person}{Shashank Subramanya}, \bibinfo{person}{Janneke van~de Pol}, {and} \bibinfo{person}{Anique de Bruin}.} \bibinfo{year}{2024}\natexlab{}.
\newblock \showarticletitle{Evaluating Sentence-BERT-powered learning analytics for automated assessment of students' causal diagrams}.
\newblock \bibinfo{journal}{\emph{Journal of Computer Assisted Learning}} \bibinfo{volume}{40}, \bibinfo{number}{6} (\bibinfo{year}{2024}), \bibinfo{pages}{2667--2680}.
\newblock


\bibitem[Ramesh and Sanampudi(2022)]%
        {ramesh2022automated}
\bibfield{author}{\bibinfo{person}{Dadi Ramesh} {and} \bibinfo{person}{Suresh~Kumar Sanampudi}.} \bibinfo{year}{2022}\natexlab{}.
\newblock \showarticletitle{An automated essay scoring systems: a systematic literature review}.
\newblock \bibinfo{journal}{\emph{Artificial Intelligence Review}} \bibinfo{volume}{55}, \bibinfo{number}{3} (\bibinfo{year}{2022}), \bibinfo{pages}{2495--2527}.
\newblock


\bibitem[Reimers and Gurevych(2019)]%
        {reimers2019sentence}
\bibfield{author}{\bibinfo{person}{Nils Reimers} {and} \bibinfo{person}{Iryna Gurevych}.} \bibinfo{year}{2019}\natexlab{}.
\newblock \showarticletitle{Sentence-bert: Sentence embeddings using siamese bert-networks}.
\newblock \bibinfo{journal}{\emph{arXiv preprint arXiv:1908.10084}} (\bibinfo{year}{2019}).
\newblock


\bibitem[Schafer et~al\mbox{.}(2005)]%
        {schafer2005resistance}
\bibfield{author}{\bibinfo{person}{William~D Schafer}, \bibinfo{person}{Phill Gagn{\'e}}, {and} \bibinfo{person}{Robert~W Lissitz}.} \bibinfo{year}{2005}\natexlab{}.
\newblock \showarticletitle{Resistance to confounding style and content in scoring constructed-response items}.
\newblock \bibinfo{journal}{\emph{Educational Measurement: Issues and Practice}} \bibinfo{volume}{24}, \bibinfo{number}{2} (\bibinfo{year}{2005}), \bibinfo{pages}{22--28}.
\newblock


\bibitem[Sha et~al\mbox{.}(2022)]%
        {sha2022bigger}
\bibfield{author}{\bibinfo{person}{Lele Sha}, \bibinfo{person}{Yuheng Li}, \bibinfo{person}{Dragan Gasevic}, {and} \bibinfo{person}{Guanliang Chen}.} \bibinfo{year}{2022}\natexlab{}.
\newblock \showarticletitle{Bigger data or fairer data?: augmenting bert via active sampling for educational text classification}. In \bibinfo{booktitle}{\emph{International Conference on Computational Linguistics 2022}}. Association for Computational Linguistics (ACL), \bibinfo{pages}{1275--1285}.
\newblock


\bibitem[Siedahmed et~al\mbox{.}(2025)]%
        {Siedahmed2025nonstandard}
\bibfield{author}{\bibinfo{person}{Abubakir Siedahmed}, \bibinfo{person}{Jaclyn Ocumpaugh}, \bibinfo{person}{Zelda Ferris}, \bibinfo{person}{Dinesh Kodwani}, \bibinfo{person}{Neil Heffernan}, {and} \bibinfo{person}{Eamon Worden}.} \bibinfo{year}{2025}\natexlab{}.
\newblock \showarticletitle{Nonstandard English and the Automated Scoring of Open-Ended Math Problems}. In \bibinfo{booktitle}{\emph{Proceedings of the 18th Intl. Conference on Educational Data Mining}}. \bibinfo{address}{Palermo, Italy}, \bibinfo{pages}{254--264}.
\newblock
\showISBNx{978-1-7336736-6-2}


\bibitem[Snijders and Bosker(2011)]%
        {snijders2011multilevel}
\bibfield{author}{\bibinfo{person}{Tom~AB Snijders} {and} \bibinfo{person}{Roel Bosker}.} \bibinfo{year}{2011}\natexlab{}.
\newblock \showarticletitle{Multilevel analysis: An introduction to basic and advanced multilevel modeling}.
\newblock  (\bibinfo{year}{2011}).
\newblock


\bibitem[S{\"u}zen et~al\mbox{.}(2020)]%
        {suzen2020automatic}
\bibfield{author}{\bibinfo{person}{Neslihan S{\"u}zen}, \bibinfo{person}{Alexander~N Gorban}, \bibinfo{person}{Jeremy Levesley}, {and} \bibinfo{person}{Evgeny~M Mirkes}.} \bibinfo{year}{2020}\natexlab{}.
\newblock \showarticletitle{Automatic short answer grading and feedback using text mining methods}.
\newblock \bibinfo{journal}{\emph{Procedia computer science}}  \bibinfo{volume}{169} (\bibinfo{year}{2020}), \bibinfo{pages}{726--743}.
\newblock


\bibitem[Xavier et~al\mbox{.}(2025)]%
        {xavier2025}
\bibfield{author}{\bibinfo{person}{Christian Xavier}, \bibinfo{person}{Lucas Rodrigues}, \bibinfo{person}{Nuno Costa}, \bibinfo{person}{Ricardo Neto}, \bibinfo{person}{Gustavo Alves}, \bibinfo{person}{Taciana~Pontual Falc{\~a}o}, \bibinfo{person}{others}, {and} \bibinfo{person}{Rafael~Ferreira Mello}.} \bibinfo{year}{2025}\natexlab{}.
\newblock \showarticletitle{Empowering Instructors With {AI}: Evaluating the Impact of an {AI}-Driven Feedback Tool in Learning Analytics}.
\newblock \bibinfo{journal}{\emph{{IEEE} Transactions on Learning Technologies}} (\bibinfo{year}{2025}).
\newblock


\bibitem[Xu and Pardos(2024)]%
        {xu2024extracting}
\bibfield{author}{\bibinfo{person}{Yinuo Xu} {and} \bibinfo{person}{Zach~A Pardos}.} \bibinfo{year}{2024}\natexlab{}.
\newblock \showarticletitle{Extracting course similarity signal using subword embeddings}. In \bibinfo{booktitle}{\emph{Proceedings of the 14th Learning Analytics and Knowledge Conference}}. \bibinfo{pages}{857--863}.
\newblock


\bibitem[Yang et~al\mbox{.}(2024)]%
        {yang2024leveraging}
\bibfield{author}{\bibinfo{person}{Kexin~Bella Yang}, \bibinfo{person}{Conrad Borchers}, \bibinfo{person}{Ann-Christin Falhs}, \bibinfo{person}{Vanessa Echeverria}, \bibinfo{person}{Shamya Karumbaiah}, \bibinfo{person}{Nikol Rummel}, {and} \bibinfo{person}{Vincent Aleven}.} \bibinfo{year}{2024}\natexlab{}.
\newblock \showarticletitle{Leveraging multimodal classroom data for teacher reflection: Teachers’ preferences, practices, and privacy considerations}. In \bibinfo{booktitle}{\emph{European Conference on Technology Enhanced Learning}}. Springer, \bibinfo{pages}{498--511}.
\newblock


\bibitem[Zhang et~al\mbox{.}(2025)]%
        {zhang2025deconfounding}
\bibfield{author}{\bibinfo{person}{Guixian Zhang}, \bibinfo{person}{Guan Yuan}, \bibinfo{person}{Debo Cheng}, \bibinfo{person}{Lin Liu}, \bibinfo{person}{Jiuyong Li}, \bibinfo{person}{Ziqi Xu}, {and} \bibinfo{person}{Shichao Zhang}.} \bibinfo{year}{2025}\natexlab{}.
\newblock \showarticletitle{Deconfounding representation learning for mitigating latent confounding effects in recommendation: G. Zhang et al.}
\newblock \bibinfo{journal}{\emph{Knowledge and Information Systems}} \bibinfo{volume}{67}, \bibinfo{number}{7} (\bibinfo{year}{2025}), \bibinfo{pages}{5999--6020}.
\newblock


\end{thebibliography}

\end{document}